\documentclass[utf8]{article}
\usepackage{graphicx}
\usepackage{jfrExamplee}
\usepackage{graphicx}
\usepackage{apalike}
\usepackage{setspace}
\usepackage{caption}
\usepackage{appendix}
\usepackage{array}
\usepackage{booktabs}
\usepackage{multirow}
\usepackage{adjustbox} 
\usepackage{tabularx} 
\usepackage{amsmath}

\title{Foundation Model-Driven Grasping of Unknown Objects via Center of Gravity Estimation}

\author{
Kang Xiangli \\
School of Electronics and Control Engineering\\
Chang’an University\\
Xi'an, Shaanxi, China\\
\texttt{kxiangli@chd.edu.cn} \\
\And
Yage He \\
Chang’an University \\
Xi'an, Shaanxi, China \\
\texttt{heyage@chd.edu.cn} \\
\AND
 Xianwu~Gong\thanks{ Correspondence: Xianwu gong, School of Electronics and Control Engineering, Chang’an University,
Xi'an, Shaanxi, China, Email: xwgong@chd.edu.cn}\\
Chang’an University \\
Xi'an, Shaanxi, China \\
\texttt{xwgong@chd.edu.cn} \\
\And
Zehan Liu \\
Chang’an University  \\
Xi'an, Shaanxi, China \\
\texttt{chdlzh@chd.edu.cn} \\
\And
Yuru Bai \\
Chang’an University \\
Xi'an, Shaanxi, China \\
\texttt{Baiyuru@chd.edu.cn} \\
}

\begin{document}

\maketitle

\begin{abstract}
This study presents a grasping method for objects with uneven mass distribution by leveraging diffusion models to localize the center of gravity (CoG) on unknown objects. In robotic grasping, CoG deviation often leads to postural instability, where existing keypoint-based or affordance-driven methods exhibit limitations. We constructed a dataset of 790 images featuring unevenly distributed objects with keypoint annotations for CoG localization. A vision-driven framework based on foundation models was developed to achieve CoG-aware grasping. Experimental evaluations across real-world scenarios demonstrate that our method achieves a 49\% higher success rate compared to conventional keypoint-based approaches and an 11\% improvement over state-of-the-art affordance-driven methods. The system exhibits strong generalization with a 76\% CoG localization accuracy on unseen objects, providing a novel solution for precise and stable grasping tasks.
\end{abstract}

\section{Introduction}

General-purpose robotic grasping is crucial for robot development \cite{hu2024generalpurposerobotsfoundationmodels}. It involves high-level task planning and low-level control. Recent studies \cite{huang2023groundeddecodingguidingtext,huang2022innermonologueembodiedreasoning} show that high-level planning relies on Visual-Language Models' (VLM) understanding, which provides explainability and reasoning, vital for general-purpose robotic grasping. For open-vocabulary grasping \cite{liu2024mokaopenworldroboticmanipulation}, VLM's common sense helps identify the same target areas on objects as humans do; this area is called affordance \cite{b870c976-99e6-3285-8c2c-ab8adb2892f7}. However, relying solely on the affordance of the grasping area is not precise enough, especially when grasping objects with uneven mass distribution.

Then, there is this problem: When grasping an object with an uneven mass distribution, the end-of-the-object may differ in size or material, leading to inconsistent gravity effects. During grasping, the object may tilt or even fail to be lifted. Even if grasping succeeds, it might fall unexpectedly during transport, potentially damaging the floor or table. For instance, a hammer has a handle and head made of different materials, experiencing different gravitational forces. According to the definition of the affordance area, the grasping area should be the handle area. If the robot grasps the handle, the hammer head may not be lifted. During transportation, this could lead to collisions with other objects. When a person grabs a hammer, they instinctively place their hand closer to the head, near the center of gravity. This grip is not only more efficient, but also reduces the risk of accidental drops. Thus, locating an object's center of gravity is essential for effective grasping.

Recent research \cite{chen2022keypoint,GUI2025111318} has shown that most grasping methods generate grasp locations from RGB-D images of the current scene. These methods can be categorized into planar 3D \cite{10517376} and parallel jaw grasping 6D \cite{wen2024foundationposeunified6dpose}, depending on the task. Planar grasping restricts objects to the work surface, allowing grippers to grasp from one direction. In contrast, 6D grasping enables grippers to grasp at any angle and position in 3D space. In this paper, we mainly focus on 6D grasping. In previous work, \cite{liu2022centerofmassbasedrobustgrasppose} to find an object's center of gravity, most approaches used tactile sensors to collect data and train a neural network for this purpose. However, this approach requires additional sensors and rarely succeeds on the first grasp. Thus, it presents a significant challenge to achieving zero-shot grasping.

Traditional vision-based grasping systems predominantly rely on geometric affordance estimation from RGB/RGB-D images to generate grasp configurations. While this paradigm achieves satisfactory performance on homogeneous objects with uniform mass distribution, its effectiveness significantly degrades when handling multi-material objects or geometrically asymmetric structures. The performance bottleneck stems from the inherent limitation of visual perception in inferring latent physical properties (e.g., center of mass). In contrast, humans instinctively integrate prior material understanding and gravitational reasoning - acquired through lifelong interaction-to predict approximate mass distribution patterns. This neurocognitive strategy enables stable grasps even for novel objects through intuitive CoG estimation. For robotic grippers with limited dexterity, explicitly incorporating such gravity-aware reasoning becomes crucial to prevent slip-induced failures during physical interaction, particularly when handling objects with non-convex geometries or heterogeneous density distributions. This observation motivates our integration of foundation models to emulate human-like material understanding and gravitational reasoning, enabling data-efficient CoG prediction for unseen objects through physical commonsense learning.

This paper uses the generalization of diffusion and foundation models to propose a new grasping framework that integrates the physical property of center of gravity (CoG) into robot manipulation. To better identify target objects, we use the advanced common sense reasoning and accurate spatial understanding of VLMs for a coarse-to-fine scene understanding. To address the grasping of objects with uneven mass distribution, we collect a dataset of such common objects from daily life and work, annotating their CoG which is determined by suspending them with a string. Based on the existing CoG data, diffusion models can generalize to unseen objects and locate their CoG effectively.

The algorithm can achieve high-successm rate grasping in real-world scenarios. By incorporating the CoG concept, it effectively prevents accidental drops and imbalance during the grasping of objects with uneven mass distribution. This improvement significantly exceed the  KGNv2 baseline model \cite{chen2023kgnv2}.

The contributions of this paper can be summarized as follows:

\begin{itemize}
\item[$\bullet$] We propose a novel framework that integrates the physical attribute of center of gravity (CoG) into the grasping process, enabling robots to grasp objects based on their CoG. This approach effectively prevents imbalance and accidental drops during grasping.
\item[$\bullet$] We construct a dataset of objects with uneven mass distribution. The CoG coordinates in this dataset are determined using the suspension method, ensuring they reflect real-world physical properties. The dataset includes CoG information for everyday items.
\item[$\bullet$] Our research demonstrates that robots can grasp objects with uneven mass distribution without additional sensors. This work provides a strong foundation for future research and advances in robotic grasping.
\end{itemize}

\section{Related Work}

\textbf{Foundational Models in Robotics.} Recent studies \cite{hu2023lookleapunveilingpower,Driess2023PaLMEAE,brohan2023rt1roboticstransformerrealworld} indicate that foundational models, particularly Visual-Language Models (VLM), are transforming robotics. They excel in environmental understanding and reasoning. In previous work, Open VLM \cite{kim24openvla}fine-tuned VLM to directly output robotic arm motion trajectories but relied on large datasets. To avoid using large datasets, SeeDo \cite{wang2024vlmseerobotdo} uses human-grasping videos as data. It leverages foundational models to generate action commands, which are then converted into executable robotic-arm code via Language Model Programs (LMPs). However, this method only achieves one-shot grasping and can't perform zero-shot grasping on unseen objects. Rekep \cite{huang2024rekep} locates an object's keypoints via a vision foundation model to generate task-execution constraint functions. However, this method depends on complex prompt engineering and only has a coarse understanding of the scene. This work leverages the strong environmental understanding of foundational models for fine-grained scene perception. It can generalize to unseen objects with simple prompt engineering.

\textbf{Task-Oriented Grasping.} Task-oriented grasping adjusts the grasping position on an object according to the specific task being performed. Recent studies  have explored using VLMs for task-oriented object grasping. For instance, OOAL \cite{li2023oneshotopenaffordancelearning}, a representative approach, enhances grasping by improving the alignment between visual features and affordance text embeddings within a visual-language framework. Robo abc \cite{ju2024roboabcaffordancegeneralizationcategories} leverages CLIP's retrieval capabilities and memory to find similar items, then generalizes to unseen objects using diffusion models. Existing methods mainly depend on current datasets. For objects with uneven mass distribution, there's a lack of CoG-related datasets. Simply learning grasp locations from existing datasets can't solve the dropping issue caused by uneven force distribution on the object's ends. To address this, this paper offers a dataset with center-of-gravity info, effectively resolving the mentioned issues.

\textbf{enter-of-Gravity Grasping.} For objects with uneven mass distribution, grasping far from the CoG may lead to failure. Recent research \cite{Wang2022CenterofMassBasedRO}has applied reinforcement learning to locate the center of gravity through trial and error in simulations. However, this approach faces the challenge of transferring findings from simulation to the real world (sim-to-real problem). Robotic CoG regrasping using RL \cite{Wang2022CenterofMassBasedRO} with tactile perception is trained in simulation, leading to sim-to-real challenges and an inability to perform zero-shot grasping. The aforementioned approaches require sensors to collect data and involve trial-and-error to determine the CoG-grasping position. In contrast, this paper employs diffusion models to locate the CoG without trial-and-error, enabling zero-shot grasping and avoiding the sim-to-real issue.

\section{Method}

In this section, we first introduce the setup of the manipulation task in Section 3.1. Next, in Section 3.2, we discuss the construction of a memory bank and the generalization to unseen objects using diffusion models. Finally, in Section 3.3, we explore task-oriented grasping.

\begin{figure}[!tbp]
    \centering
    \includegraphics[width=1\textwidth]{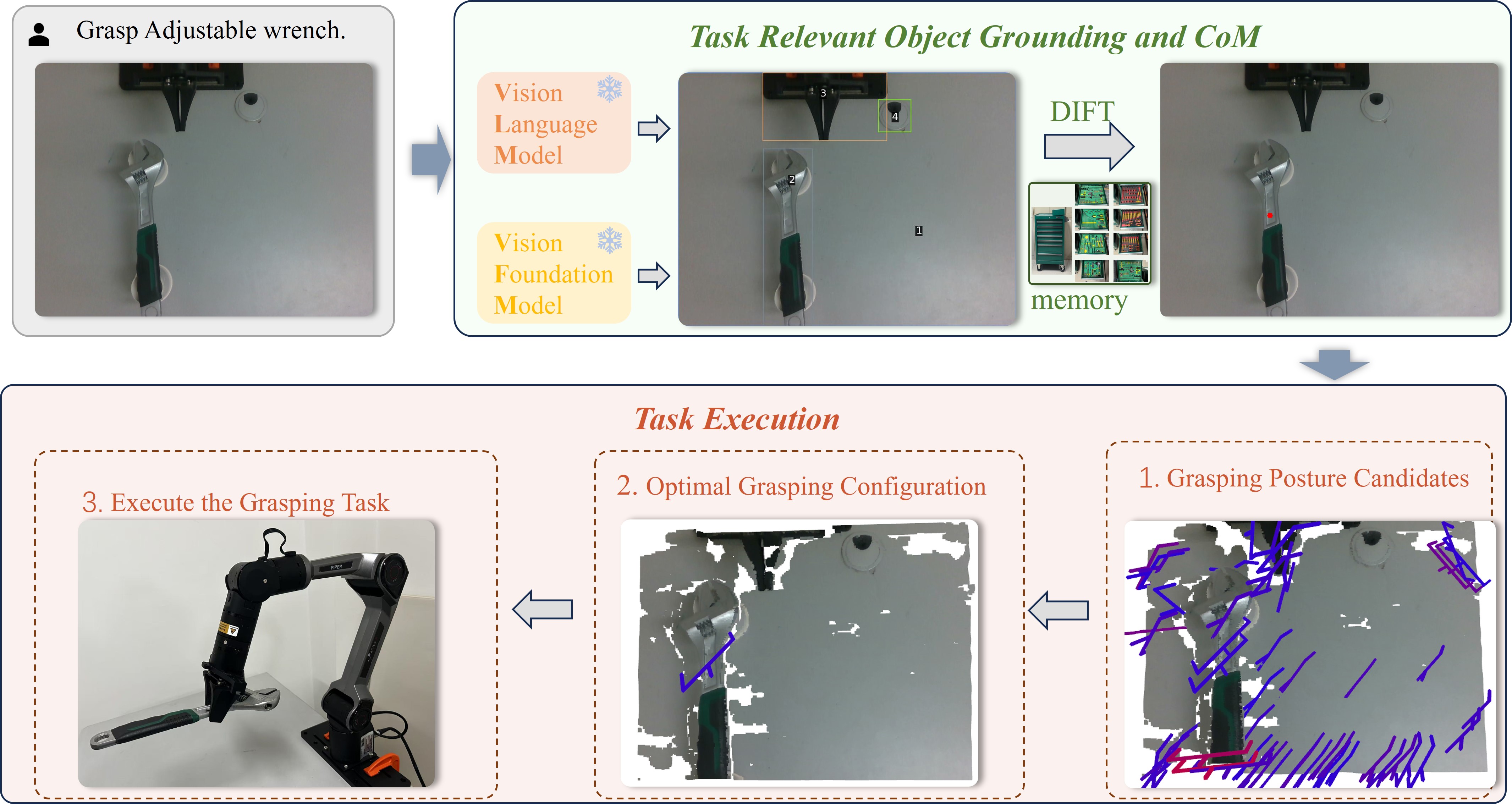} 
    \caption{Overview framework.Given the instruction and RGB-D images, the VLM first identifies objects relevant to the experiment. It then retrieves similar images from memory and maps them to the target object using DIFT. Finally, it selects the optimal grasping pose.}
    \label{fig:mian}
\end{figure}

\subsection{Promblem Formulation}

Task-oriented grasping consists of two key steps: locating the target object and performing the grasp. For example, grasping a hammer requires first identifying it and then determining its center of gravity (CoG) for a stable grip. Grasping near the CoG ensures efficient force application and prevents accidental drops. Based on this, our method comprises two modules: the Target Object Localization Module and the CoG Grasping Module.

\begin{figure}[!tbp]
    \centering
    \includegraphics[width=1\textwidth]{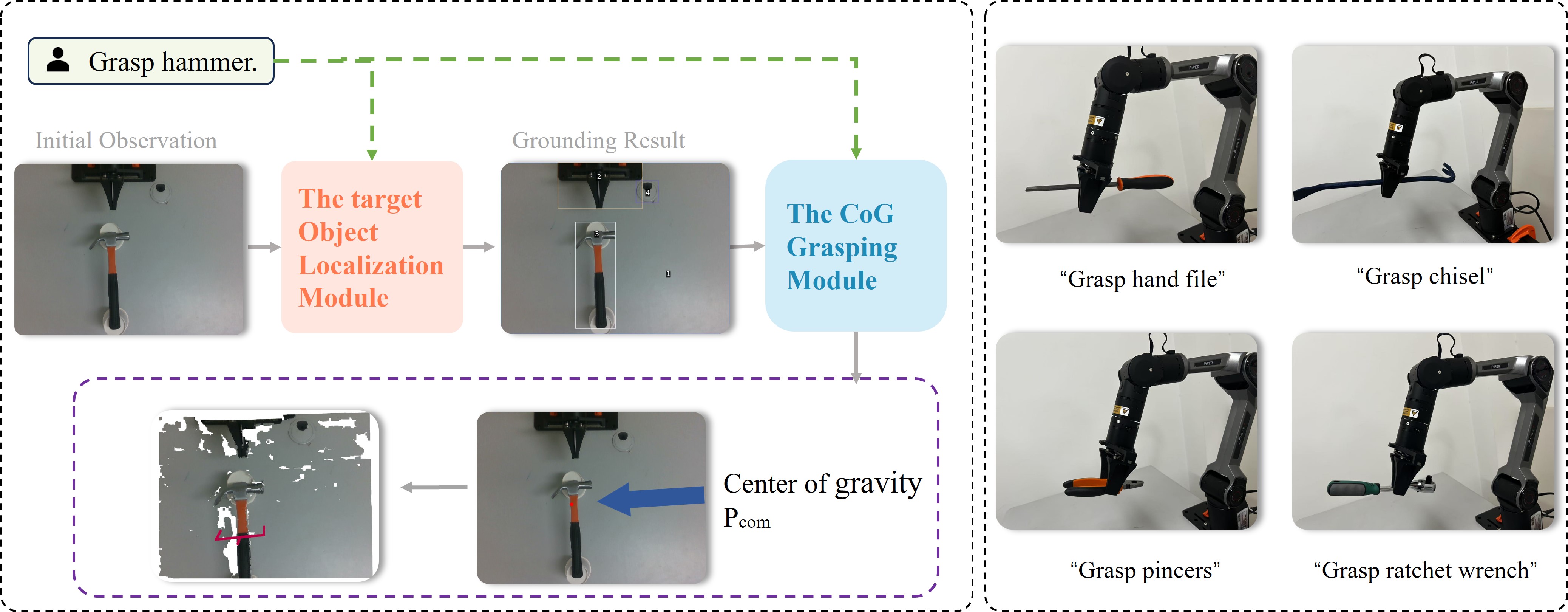} 
    \caption{\textbf{left.} Given the instruction and current RGB-D images, the VLM first identifies the target object. It then generalizes the segmented image and maps the center of gravity to the current scene using the center-of-gravity module, selecting the nearest and highest-scoring grasp pose. \textbf{right.} The right figure shows an example of a real-world experiment. The method can be extended to open-world scenarios, enabling parallel grasping of target objects without additional training.}
    \label{fig:kuangjia}
\end{figure}

Given a manipulation task \( l \), we use a depth camera to capture the scene information \( O \) (RGB-D).  A vision foundation model SAM \cite{kirillov2023segment} segments and labels all objects in the scene. With the help of a VLM's commonsense reasoning, we filter out irrelevant objects, obtaining the target object's image \( I_{{target}} \). This completes the Target Object Localization Module( Figure \ref{fig:kuangjia}).

Next, the Center-of-Gravity (CoG) Grasping Module is executed. We build a CoG dataset \( D_0 \) for common objects. Using the generalization ability of diffusion models, we map the CoG to the target object, expressed as \( P_{target} = f(l, D) \). This point is used to filter grasping poses generated by GraspNet \cite{fang2020graspnet}, and the robotic arm performs the grasp.

\subsection{Center of Gravity (CoG) Localization}

The contact points between humans and objects are defined as affordances. When interacting with objects, humans contact different areas based on the object's shape and function. However, robotic end effectors lack the flexibility of a human hand and have difficulty identifying contact points. Recent research has relied on existing datasets that often fail to meet specific research needs. For example, when grasping an axe, existing datasets indicate interaction areas on the handle. However, the smooth wooden texture of the axe handle makes it difficult for robotic end effectors with small torque to maintain a firm grasp. Additionally, the axe's head and handle, made of different materials, experience uneven gravitational forces. This imbalance can lead to accidental slippage during grasping.

\subsubsection{Data Collection}

To solve this, we collect common workplace items, using two CoG-determination methods: suspension and SAM segmentation. Firstly, we gather numerous images of daily-life items like hammers, wrenches, and screwdrivers. Two types of images are collected: one with the item alone, and another with the item suspended by a rope. \textbf{Suspension Method:} Suspend the item to locate its balance point, then label this point on the item. \textbf{SAM Segmentation:} Acquire the contact position from the suspended-item image, map it to the item-alone image as a region, and determine the CoG as the region's center. Both methods yield CoG data with real-world physical properties.

\subsubsection{Memory Construction and CoG Mapping}

After data collection, we use the CLIP image encoder to map each source image \( I_s \) to a feature vector, which is stored as memory. When the target image \( I_t \) is obtained, the CLIP image encoder maps it to another feature vector. We then calculate the cosine similarity between these two feature vectors and select the three images with the highest similarity as the source images \( I_{diff} \) for the diffusion model.

This process is formulated as:

\begin{equation} \label{eq:similarity}
I_{\text{diff}} = \text{topk} \left\{ \text{similarity} \right\} = \text{topk} \left\{ \cos\left(\text{CLIP}_v(I_s), \text{CLIP}_v(I_t)\right) \right\}, \quad k=3
\end{equation}
After retrieving the most similar objects and CoG points from memory, this paper uses semantic correspondence to map these features to the current scene and object. Semantic correspondence maps points from the source image to the target image. In this work, we utilize the feature extraction, implicit learning, and cross - modal alignment capabilities of the diffusion model DIFT \cite{tang2023emergent} to guide robotic manipulation in unfamiliar environments. More specifically, given the source image \(I_{{diff}}\), target image \(I_t\), and source point \(p_{{diff}}\), our goal is to find the corresponding point \(p_t\) in the target image.

To enhance the accuracy of locating an object's center of gravity (CoG), this paper combines diffusion models with a visual-language model (VLM). Diffusion models can generate candidate CoG points but may produce mismatches between source and target images. VLMs, while unable to directly output precise CoG locations, excel in selecting the most plausible point using their commonsense reasoning. Our approach leverages diffusion models to generate multiple candidate points and then employs VLMs to choose the point closest to the actual CoG, ensuring accurate localization in unfamiliar environments.

\subsubsection{Task-Oriented Grasping}

This paper uses the trained GraspNet to generate grasping poses. GraspNet, trained on massive datasets, contains over one billion grasping poses. The process starts with acquiring RGB-D information via a depth camera, which is converted into point clouds and backprojected into the camera's 3D space. These point clouds serve as inputs for GraspNet, which then outputs 6-DOF grasping poses. The output includes details like grasp points, width, height, depth, and a "grasp score" indicating the likelihood of a successful grasp.
However, selecting only the highest scoring pose often fails to meet task requirements. Therefore, a crucial filtering step is necessary. This filtering is based on specific task needs, such as proximity to the object's center of gravity or avoidance of collisions with other items during transportation. Considering these factors, the filtering process refines the grasping pose options to better suit the particular task at hand.

This paper designs a coarse-to-fine filtering process. First, SAM segments the image into regions with unique identifiers. The image and action commands are input into a VLM, which outputs the target object's identifier. This completes the first-layer extraction. Then, the CoG mapping module projects the CoG coordinates onto the target object image. GraspNet generates grasping poses. To filter these poses in the plane, they are projected onto the image, and the one closest to the CoG is selected, with rotation correction performed. After these two steps, the most appropriate grasp position is found accurately. Finally, the grasp pose with the highest GraspNet score and closest to the CoG is chosen as the target, and the robotic arm executes the operation.

To address the issue of object movement invalidating previously generated grasping poses, this paper implements a closed-loop operation. Before executing the grasping command, it checks if the current pose's projected coordinates match the previous ones. If they don't match, it re-executes the localization and CoG grasping modules. If they do match, the grasping operation proceeds. This simple design effectively handles dynamic scene changes.

\begin{figure}[!tbp]
    \centering
    \includegraphics[width=1\textwidth]{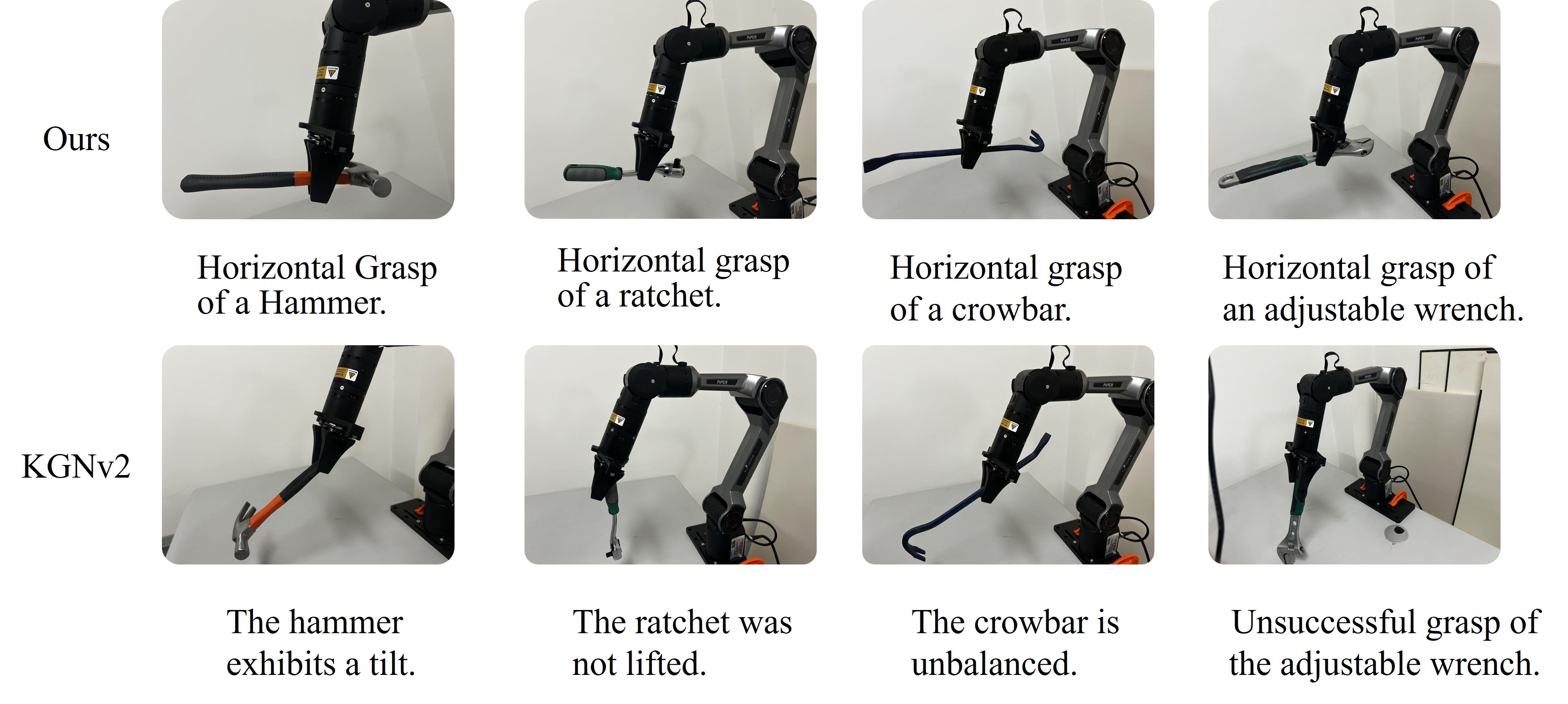} 
    \caption{\textbf{Comparison with KGNv2.}  The top row shows the results of our proposed algorithm, while the bottom row shows the results of the comparison algorithm. Our method demonstrates the ability to accurately locate the center of mass and perform parallel grasping for unseen objects. The objects from left to right are hammer, ratchet wrench, crowbar, and adjustable wrench.}
    \label{fig:duibi}
\end{figure}

\section{Experiments}

In Section 4, we first introduce the experimental setup. Then, in Section 4.2, we evaluate the real-world performance of our method. In Section 4.3, we highlight the importance of CoG integration and compare it with the baseline model KGNv2. Finally, in Section 4.4, we conduct ablation studies to analyze the key components of our framework.

\subsection{Experimental Setup}

\textbf{Hardware.} In our real-world experiments, we use a 6-DOF Piper manipulator from Sunling and a 1-DOF parallel jaw. For perception, we employ a calibrated RealSense D435 RGB-D camera.

\textbf{Tasks and Evaluation.} This paper designs 10 real-world grasping tasks, each requiring a comprehensive understanding of the target object's physical properties. For each task, under the same torque, the number of successful grasps is used as the evaluation metric.

\textbf{VLM and Prompting.} This paper uses GPT-4o \cite{openai2024gpt4ocard} from OpenAI API as the LLM. Our algorithm uses minimal few-shot prompts to help the VLM understand its role.

\textbf{Baselines.} Our algorithm is compared with KGNv2, a keypoint-based grasping method that generates grasping poses for keypoints.

\subsection{Real-World Manipulation}

Our framework's real-world applicability is tested across ten object categories. 
 Table \ref{tab:jieguo} presents the quantitative results, showing an average success rate of 80\% across these categories, outperforming the KGNv2 baseline and ablation variants. The algorithm's success is attributed to its use of VLM's commonsense knowledge for fine-grained physical understanding and its CoG module. This module accurately identifies the most similar item in memory and generalizes via diffusion models to select the optimal CoG position on the target object. During the experiment, it was observed that the mapping performance was suboptimal when identifying the center of the hammer. This issue arises because diffusion models tend to prioritize color and shape features. Specifically, both the handle and the striking head of the hammer are silver and circular, leading to ambiguity in the mapping process. However, it was also noted that when dealing with a dataset containing only 9 images of a chisel, the model successfully identified the center of mass for the target object (crowbar), demonstrating strong generalization capabilities.

\subsection{Key Factors of the Algorithm}

In this section, we delve into our algorithm, which leverages common sense knowledge embedded in foundational models to locate the object CoG. Our algorithm shows significant advantages in several aspects.

\begin{table}[!tbp]
    \centering
    \caption{\textbf{Quantitative results in real-world experiments.} Our algorithm achieves a significantly higher success rate in grasping ten objects, surpassing the baseline KGNv2 and affordance-based grasping. Additionally, we conducted an ablation study to demonstrate the importance of the base model and the CoG module in our approach.}
    \label{tab:jieguo}
    \begin{tabularx}{\textwidth}{cccccc}
        \toprule
        \textbf{Tasks} & \textbf{Ours} & \textbf{KGNv2} & \textbf{Affordance} & \textbf{Ours (w/o CoG)} & \textbf{Ours (w/o VLM)} \\
        \midrule
        \centering Hammer & 40\% & 20\% & 30\% & 40\% & 10\% \\
        \midrule
        \centering Wrench & 50\% & 20\% & 40\% & 20\% & 20\% \\
        \midrule
        \centering Pincers & 80\% & 30\% & 60\% & 50\% & 30\% \\
        \midrule
        \centering T-type Allen Wrench & 90\% & 20\% & 80\% & 70\% & 0\% \\
        \midrule
        \centering Hand file & 90\% & 40\% & 90\% & 60\% & 20\% \\
        \midrule
        \centering Allen key & 80\% & 30\% & 80\% & 70\% & 0\% \\
        \midrule
        \centering Adjustable wrench & 60\% & 10\% & 30\% & 40\% & 0\% \\
        \midrule
        \centering Ratchet wrench & 90\% & 40\% & 70\% & 70\% & 0\% \\
        \midrule
        \centering Screwdriver & 90\% & 40\% & 90\% & 80\% & 30\% \\
        \midrule
        \centering Chisel & 90\% & 20\% & 80\% & 80\% & 10\% \\
        \midrule
        \multicolumn{1}{c}{\textbf{Total}} & \textbf{76\%} & \textbf{27\%} & \textbf{65\%} & \textbf{58\%} & \textbf{12\%} \\
        \bottomrule
    \end{tabularx}
\end{table}

\textbf{The Target Object Localization Module.} Many manipulation tasks require subtle physical understanding of scenes, needing fine-grained object-part recognition and comprehension of their complex attributes. Our algorithm excels here. It uses a coarse-to-fine part-grounding module to select graspable/task-relevant object parts.

\textbf{The Center of Gravity (CoG) Module.} Grasping objects with uneven mass distribution requires a grip near the CoG; otherwise, the object may drop. Our algorithm addresses this by creating a relevant dataset and employing diffusion models to accurately locate the object's CoG. It then uses a VLM's commonsense reasoning to identify the point closest to the CoG. In contrast, Rekep \cite{huang2024rekep} uses semantic clustering to extract keypoints but lacks consideration of physical properties and task awareness. Copa \cite{ju2024roboabcaffordancegeneralizationcategories}uses pixel segmentation for keypart extraction, which is semantically oriented but devoid of physical properties. In contrast, our algorithm incorporates both semantic tasks and physical properties, ensuring grasping poses align with human habits. For instance, in grasping an adjustable wrench, Copa focuses on the handle, making parallel grasping hard for low-torque robotic arms. Our algorithm, however, targets the wrench's CoG, enabling effective parallel grasping.

\textbf{Simple Prompt Engineering.} Our algorithm shows excellent generalization in diverse scenarios through simple prompt engineering. In our experiments, whether locating target objects or finding CoG coordinates from candidates, simple prompts suffice to complete the tasks.

\textbf{Closed-Loop Execution.} Even with the optimal grasp pose, open-loop execution can lead to task failure. The gripper's pose relative to the object changes during interaction. In open-loop operation, the actuator doesn't adjust, causing failure. In contrast, closed-loop operation regenerates grasp poses based on object changes, enabling successful adaptation. Compared to KGNv2's open-loop design, our algorithm is better suited for dynamic tasks.

\subsection{Ablation Study}

To verify the rationality of our framework, we carry out a series of ablation experiments in this paper.

\textbf{Visual-Language Models (VLMs).} In this ablation study, we remove the foundational visual-language model from our framework. Instead, we employ the open-vocabulary detector OV-DINO \cite{wang2024ovdinounifiedopenvocabularydetection} to locate the target object region. The highest scoring grasp pose within this region is selected for execution. If the target object isn't detected, the task is deemed failed. The evaluation is based on successful grasps.

\textbf{Center of Gravity (CoG) Module.} In this ablation study, we eliminate the CoG-screening design in our CoG module. Instead of selecting grasping poses based on proximity to the CoG, we choose the highest - scoring grasp pose within the target region. Results show this change reduces performance. For instance, when grasping a heavy adjustable wrench, not targeting the CoG leads to low success rates.

\textbf{Affordance.} This paper compares affordance-based grasping with CoG - based grasping. Experiments show CoG - based grasping has a higher success rate for objects with uneven mass distribution. This is because such objects have unbalanced forces on their ends. Grasping only one end makes it hard to maintain balance during the grasp.

\section{Discussion \& Limitations}
This paper presents a dataset with physical CoG attributes, covering ten common objects with uneven mass distribution. It enables generalization to unseen objects via diffusion models. By analyzing existing grasping frameworks, we found low success rates in grasping such objects. Consequently, we developed a CoG-integrated visual-language framework. It requires simple prompt engineering, needs no training, and extends to open-world scenarios.

However, this study also has certain limitations. The dataset collected in this paper contains only ten categories of items, and its generalization is limited to the scope of similar items. It is relatively difficult to accurately locate the center of gravity for items not included in the dataset, which somewhat limits the broad application prospects of the framework.

\section*{Acknowledgement}
 The Shaanxi Qinchuangyuan "Scientist + Engineer" Team Building Project (2022KXJ-022)
 
\bibliographystyle{apalike}
\bibliography{jfrExampleRefs}

\newpage
\appendix

\section*{Appendix}

\section{Experimental Setup}

\begin{figure}[htbp]
    \centering
    \includegraphics[width=1\textwidth]{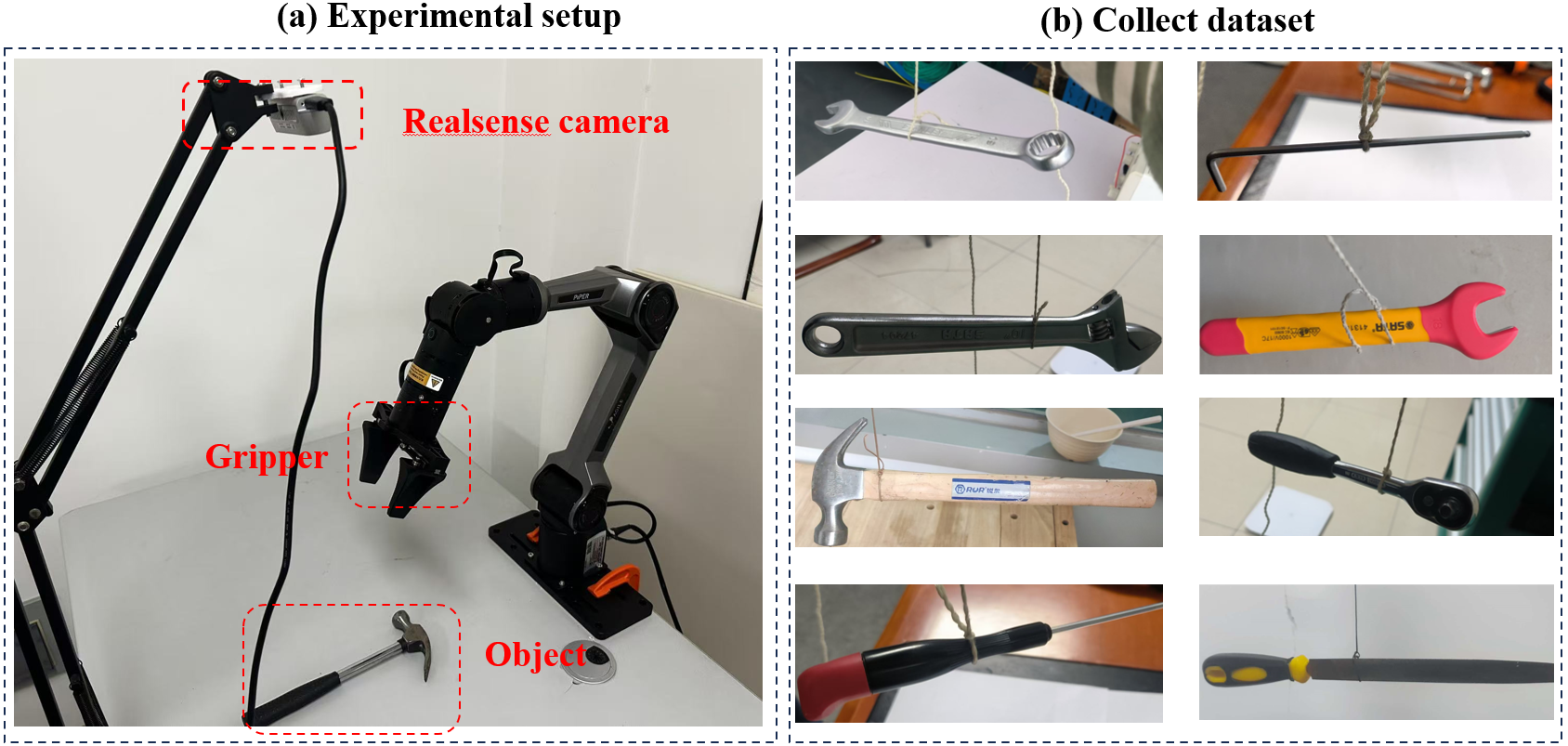} 
    \caption{\textbf{(a)}: Depicts the experimental setup, which includes a robotic arm, a depth camera, and target objects.\textbf{(b)}: Shows the hanging method used for data collection in this paper.}
    \label{fig:setup}
\end{figure}

This paper sets up a real-world desktop environment. An Agilex Piper robotic arm (6-DOF) with a 1-DOF parallel-jaw gripper is used. ROS and MoveIt1 control the robot, with RRT-Connect as the default motion planner. For perception, a calibrated Intel RealSense D435 RGB-D camera is installed above the scene. As shown in the figure \ref{fig:setup}, the left side illustrates the hardware setup of this paper.

\section{Dateset}

This paper collects 790 photos of the most common unevenly mass-distributed items in daily scenes, covering ten categories as shown in Table \ref{tab:tools_dataset}. The center of gravity for each item is determined via the suspension method and annotated.

\begin{table}[h]
    \centering
    \caption{Dataset}
    \label{tab:tools_dataset}
    \begin{tabular}{ccc ccc}
    \toprule
    \textbf{Tools} & \textbf{Images} & \textbf{Quantity} & \textbf{Tools} & \textbf{Images} & \textbf{Quantity} \\
    \midrule
    Hammer & \adjustbox{valign=c}{\includegraphics[width=0.15\textwidth]{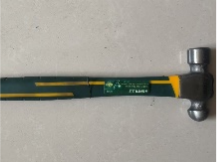}} & 75 & 
    Adjustable wrench & \adjustbox{valign=c}{\includegraphics[width=0.15\textwidth]{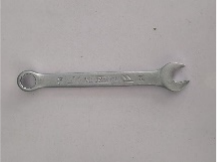}} & 54 \\
    \midrule
    Wrench & \adjustbox{valign=c}{\includegraphics[width=0.15\textwidth]{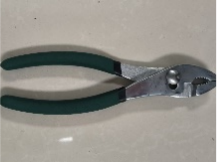}} & 241 & 
    Ratchet wrench & \adjustbox{valign=c}{\includegraphics[width=0.15\textwidth]{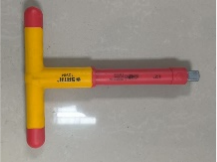}} & 29 \\
    \midrule
    Pincers & \adjustbox{valign=c}{\includegraphics[width=0.15\textwidth]{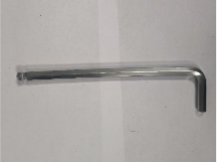}} & 60 & 
    Hand file & \adjustbox{valign=c}{\includegraphics[width=0.15\textwidth]{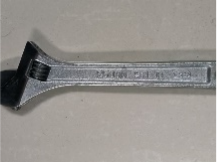}} & 76 \\
    \midrule
    T-type Allen wrench & \adjustbox{valign=c}{\includegraphics[width=0.15\textwidth]{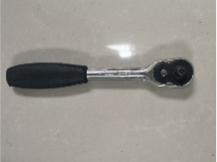}} & 44 & 
    Screwdriver & \adjustbox{valign=c}{\includegraphics[width=0.15\textwidth]{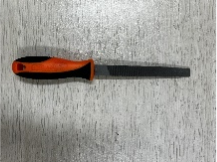}} & 114 \\
    \midrule
    Allen key & \adjustbox{valign=c}{\includegraphics[width=0.15\textwidth]{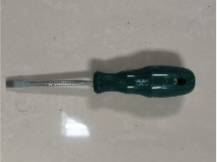}} & 94 & 
    Chisel & \adjustbox{valign=c}{\includegraphics[width=0.15\textwidth]{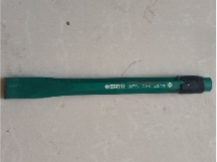}} & 9 \\
    \midrule
    \multicolumn{1}{c}{\textbf{Total}} & \multicolumn{5}{c}{\textbf{790}} \\
    \bottomrule
\end{tabular}
\end{table}

\section{Tasks}

We design 10 real - world manipulation tasks, each requiring a comprehensive understanding of the target object's physical properties. Detailed descriptions of these tasks are provided in Table II, as shown in Table \ref{tab:tasks_description}.

\section{baselines}

Our algorithm is compared with KGN, which generates grasping poses based on keypoints without extra training via foundational models. To enhance rationality, we add semantic segmentation from SAM to KGN, filtering out irrelevant keypoints. The highest - scoring point from the remaining ones becomes the target, and the robotic arm executes the grasp based on the generated pose.

\section{VLMs and prompting}
Our approach employs GPT-4o from the OpenAI API as the VLM. To facilitate the VLM's comprehension of its role, we utilize minimal few-shot prompts, which are detailed in Table 4.


\begin{table}[h]
    \centering
    \caption{Tasks description}
    \label{tab:tasks_description}
    \begin{tabularx}{\textwidth}{cX}
        \toprule
        \multicolumn{1}{c}{\textbf{Tasks}} & \multicolumn{1}{c}{\textbf{Description}} \\
        \midrule
        \centering Grasp the Hammer & The hammer is a non-uniform mass distribution object. Due to the different weights at each end, it tends to be unstable during grasping. The heavier head and lighter handle create an uneven gravitational pull, making it challenging to grasp without tilting. \\
        \midrule
        \centering Grasp the Wrench & The wrench has a non-uniform mass distribution, making it prone to instability during grasping. The varying thickness along its length results in different gravitational forces at each end, causing it to tilt if not grasped carefully. \\
        \midrule
        \centering Grasp the Pincers & Pincers are non-uniform mass distribution objects. The difference in weight between the gripping ends and the handle leads to uneven gravitational forces, causing instability during grasping. \\
        \midrule
        \centering Grasp the T-type Allen wrench & The T-type Allen wrench has a non-uniform mass distribution. The thicker T-shaped handle and thinner shaft create an imbalance in weight, leading to instability during grasping due to differing gravitational forces at each end. \\
        \midrule
        \centering Grasp the Allen key & The Allen key is a non-uniform mass distribution object. Its L-shaped or T-shaped design results in different weights at each end, causing it to be unstable during grasping as the gravitational forces are not evenly distributed. \\
        \midrule
        \centering Grasp the Adjustable wrench & The adjustable wrench has a non-uniform mass distribution. The adjustable jaw and solid handle create an imbalance in weight, leading to instability during grasping due to uneven gravitational forces. \\
        \midrule
        \centering Grasp the Ratchet wrench & The ratchet wrench is a non-uniform mass distribution object. The mechanism at one end and the handle at the other create a weight imbalance, causing it to be unstable during grasping as gravitational forces differ at each end. \\
        \midrule
        \centering Grasp the Hand file & The hand file has a non-uniform mass distribution. The dense, toothed surface and lighter handle result in different weights at each end, leading to instability during grasping due to uneven gravitational pull. \\
        \midrule
        \centering Grasp the Screwdriver & The screwdriver is a non-uniform mass distribution object. The weighted tip and lighter handle create an imbalance in weight, causing it to tilt during grasping as gravitational forces are not evenly distributed. \\
        \midrule
        \centering Grasp the Chisel & The chisel has a non-uniform mass distribution. The sharpened edge and heavier body result in different weights at each end, leading to instability during grasping due to uneven gravitational forces. \\
        \bottomrule
    \end{tabularx}
\end{table}

\section{More Visualization}

\begin{figure}[h]
	\centering
	\begin{minipage}{0.32\linewidth}
		\centering
            \textbf{Environment Image} 
		\includegraphics[width=0.9\linewidth]{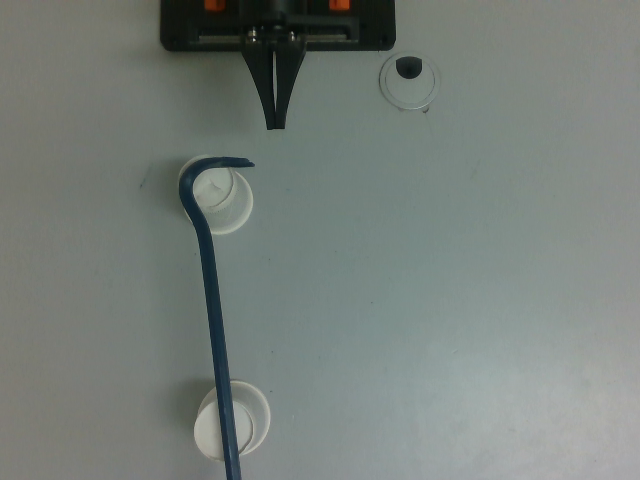}
	\end{minipage}
	\begin{minipage}{0.32\linewidth}
		\centering
            \textbf{the Localization Module} 
		\includegraphics[width=0.9\linewidth]{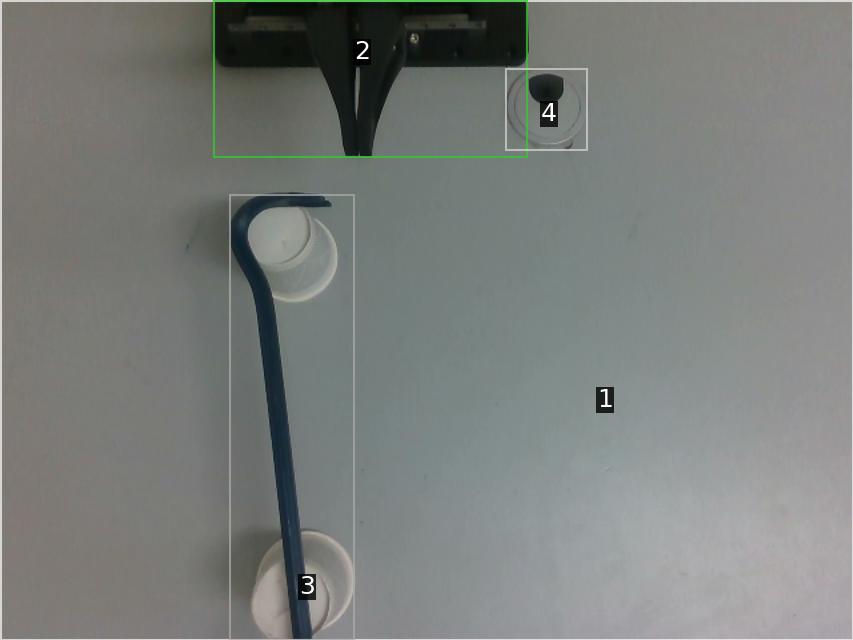}
	\end{minipage}
        \begin{minipage}{0.32\linewidth}
            \centering
            \textbf{the CoG Grasping Module}
		\includegraphics[width=0.9\linewidth]{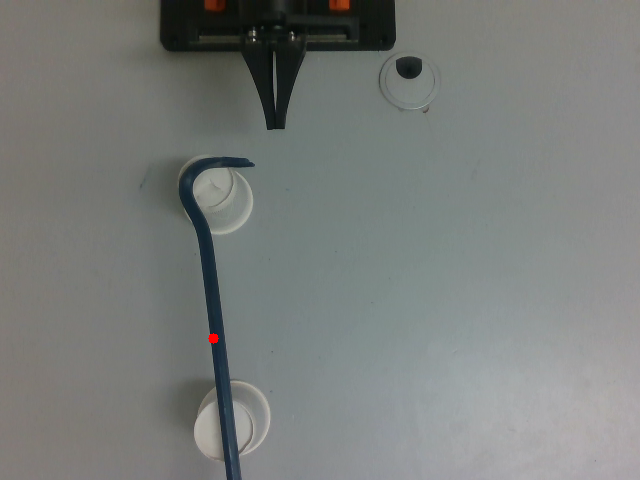}
	\end{minipage}
        \begin{minipage}{0.32\linewidth}
		\centering
		\includegraphics[width=0.9\linewidth]{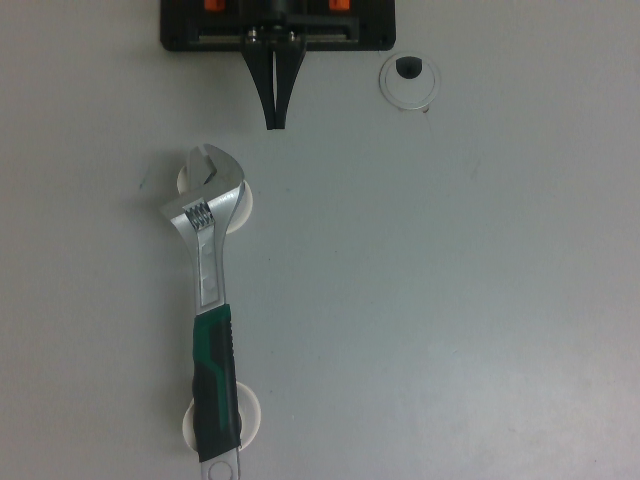}
	\end{minipage}
	\begin{minipage}{0.32\linewidth}
		\centering
		\includegraphics[width=0.9\linewidth]{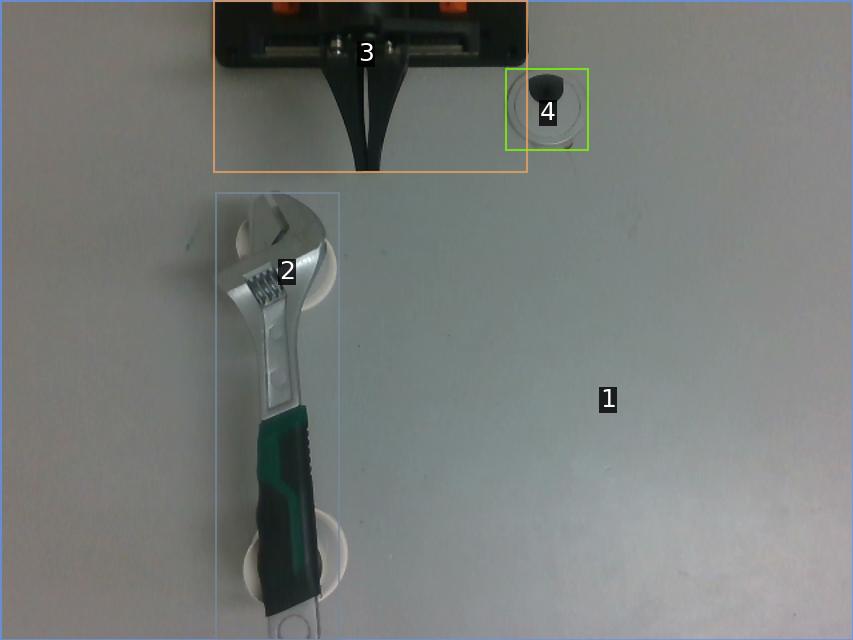}
	\end{minipage}
        \begin{minipage}{0.32\linewidth}
            \centering
		\includegraphics[width=0.9\linewidth]{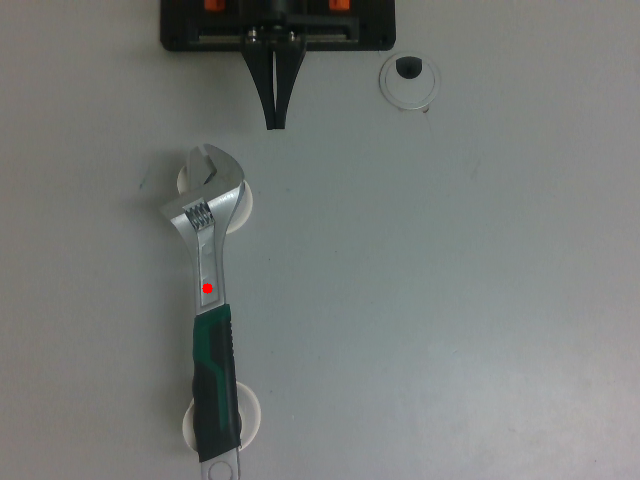}
	\end{minipage}
        \begin{minipage}{0.32\linewidth}
		\centering
		\includegraphics[width=0.9\linewidth]{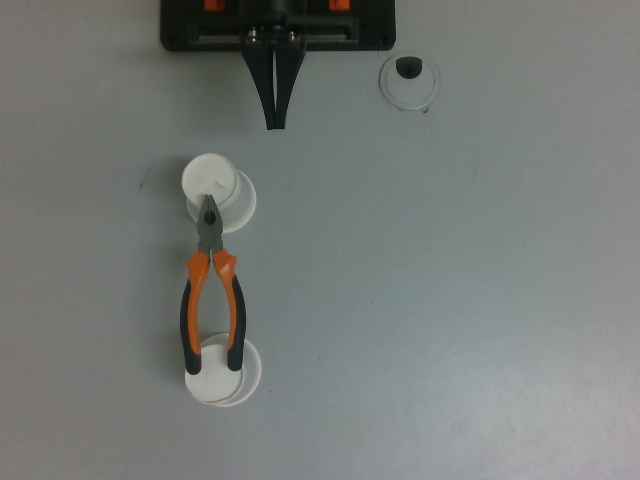}
	\end{minipage}
	\begin{minipage}{0.32\linewidth}
		\centering
		\includegraphics[width=0.9\linewidth]{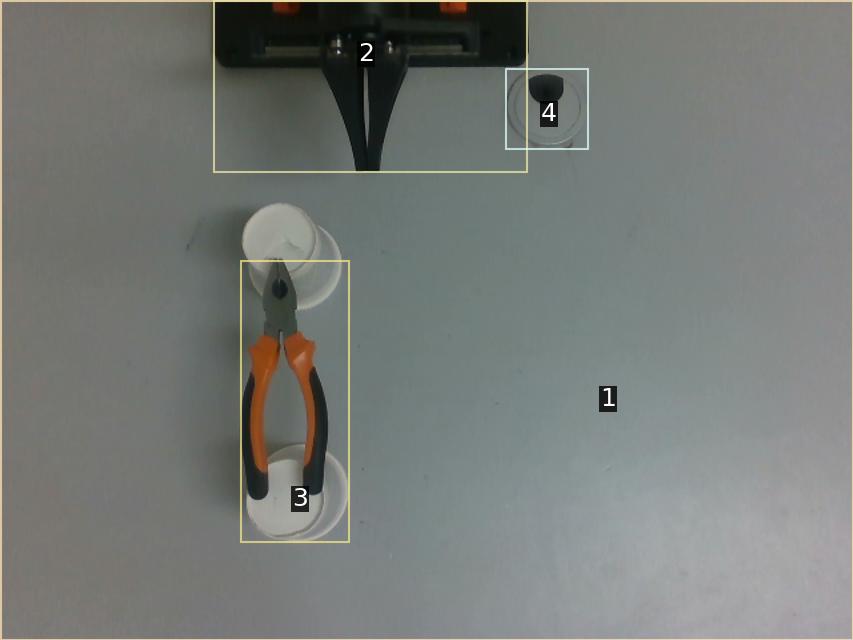}
	\end{minipage}
        \begin{minipage}{0.32\linewidth}
            \centering
		\includegraphics[width=0.9\linewidth]{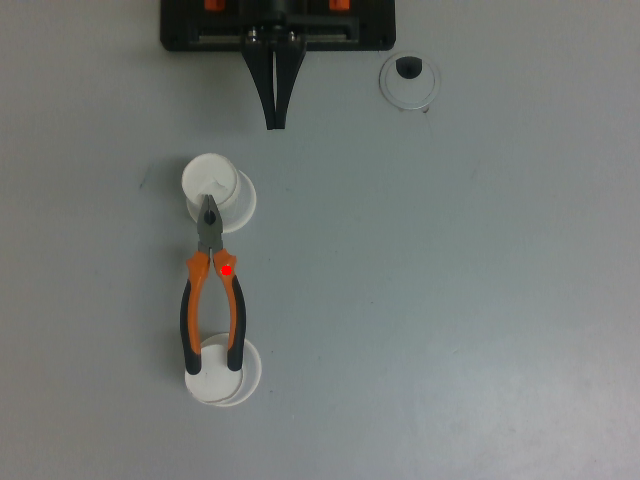}
	\end{minipage}
        \begin{minipage}{0.32\linewidth}
		\centering
		\includegraphics[width=0.9\linewidth]{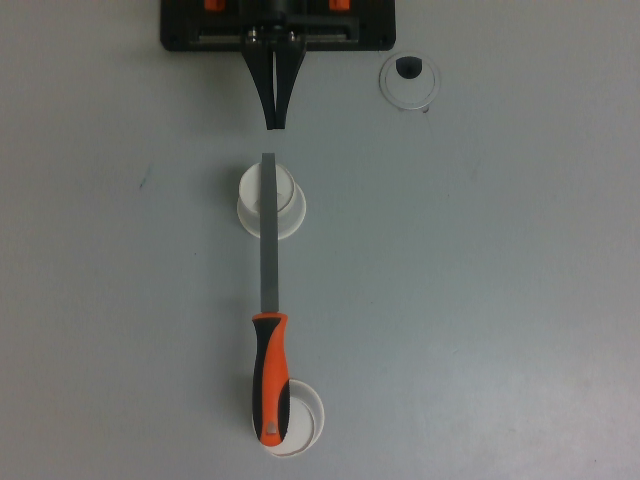}
	\end{minipage}
	\begin{minipage}{0.32\linewidth}
		\centering
		\includegraphics[width=0.9\linewidth]{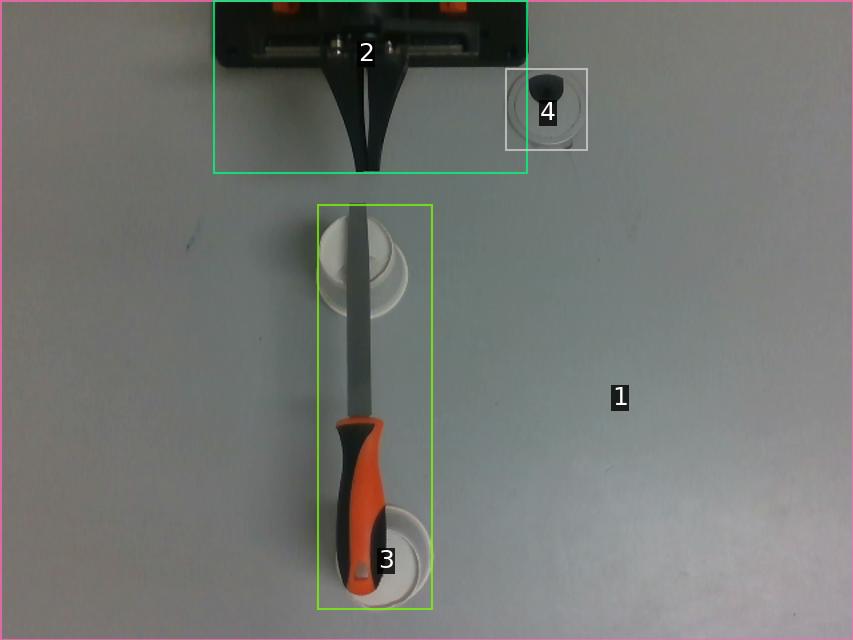}
	\end{minipage}
        \begin{minipage}{0.32\linewidth}
            \centering
		\includegraphics[width=0.9\linewidth]{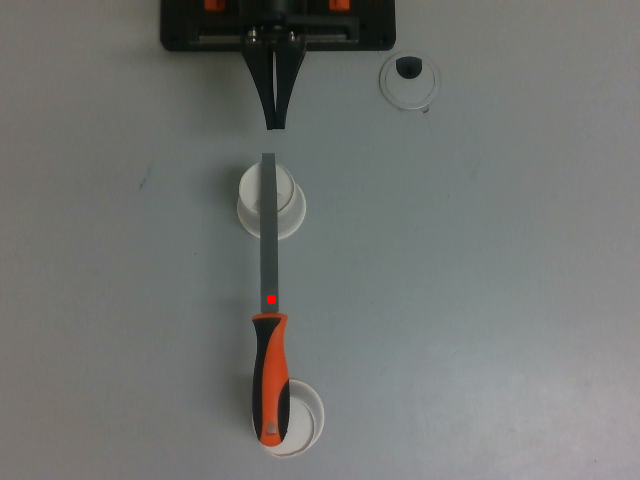}
	\end{minipage}
        \begin{minipage}{0.32\linewidth}
		\centering
		\includegraphics[width=0.9\linewidth]{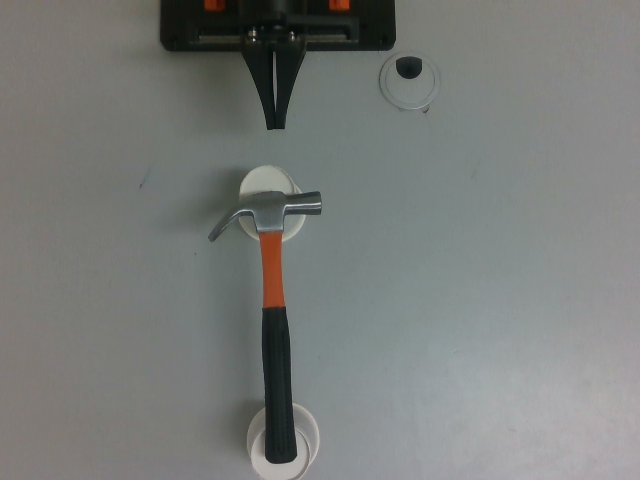}
	\end{minipage}
	\begin{minipage}{0.32\linewidth}
		\centering
		\includegraphics[width=0.9\linewidth]{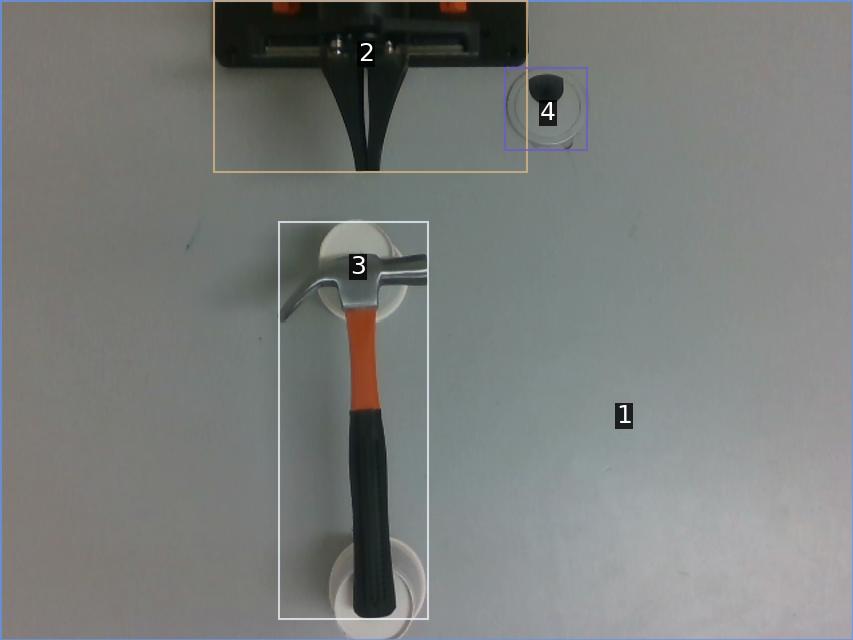}
	\end{minipage}
        \begin{minipage}{0.32\linewidth}
            \centering
		\includegraphics[width=0.9\linewidth]{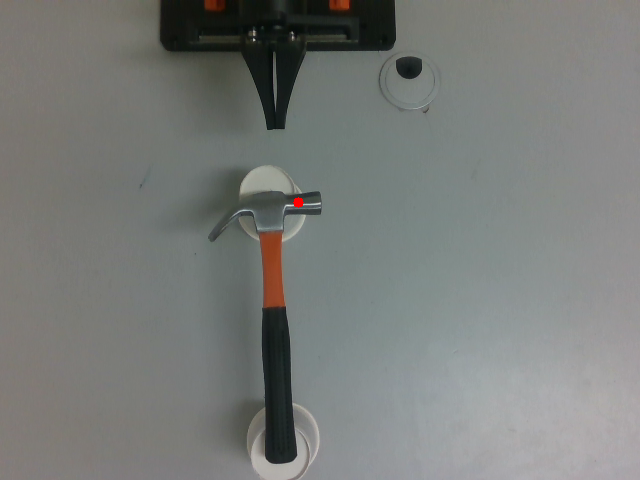}
	\end{minipage}
        \begin{minipage}{0.32\linewidth}
		\centering
		\includegraphics[width=0.9\linewidth]{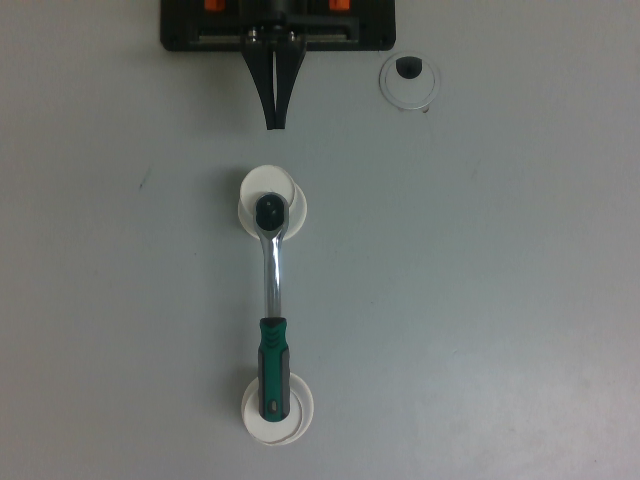}
	\end{minipage}
	\begin{minipage}{0.32\linewidth}
		\centering
		\includegraphics[width=0.9\linewidth]{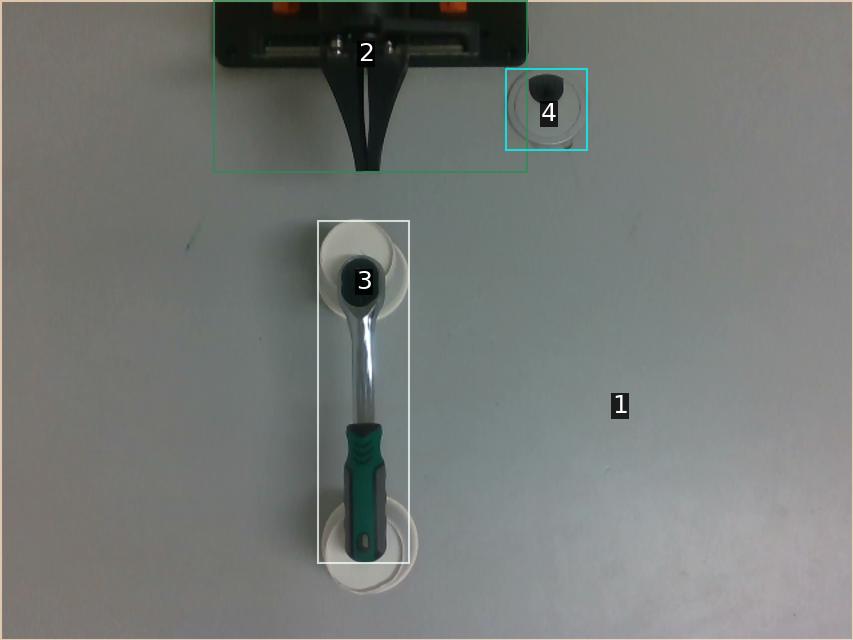}
	\end{minipage}
        \begin{minipage}{0.32\linewidth}
            \centering
		\includegraphics[width=0.9\linewidth]{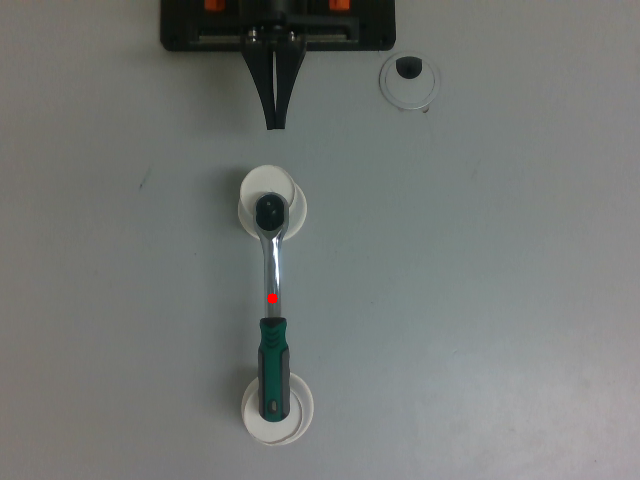}
	\end{minipage}
    \end{figure}

\begin{figure}[h]
    \centering
        \begin{minipage}{0.32\linewidth}
		\centering
		\includegraphics[width=0.9\linewidth]{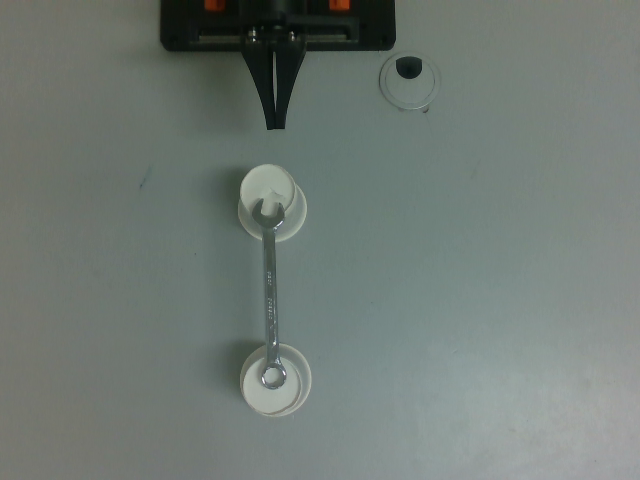}
	\end{minipage}
	\begin{minipage}{0.32\linewidth}
		\centering
		\includegraphics[width=0.9\linewidth]{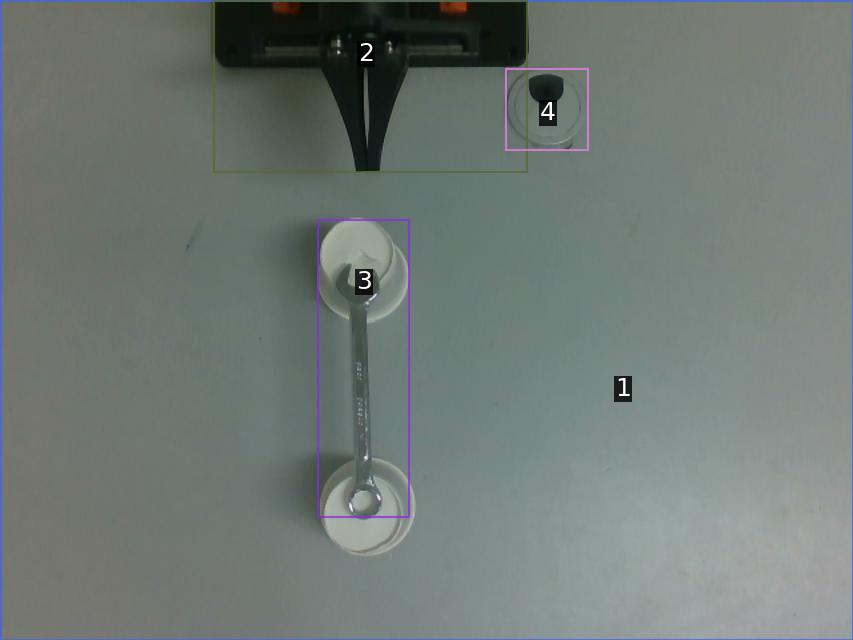}
	\end{minipage}
        \begin{minipage}{0.32\linewidth}
            \centering
		\includegraphics[width=0.9\linewidth]{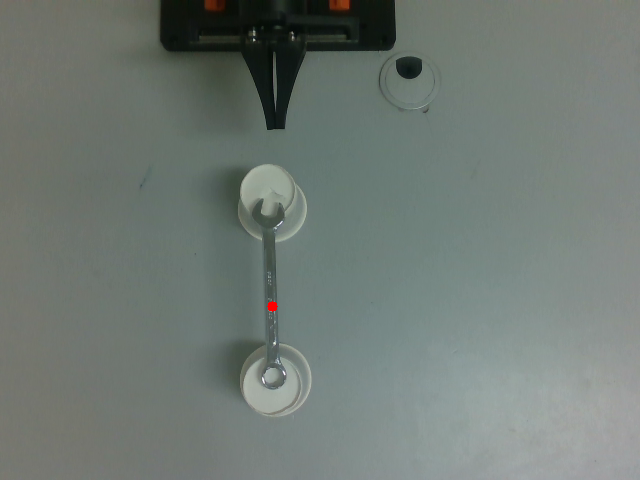}
	\end{minipage}
        \begin{minipage}{0.32\linewidth}
		\centering
		\includegraphics[width=0.9\linewidth]{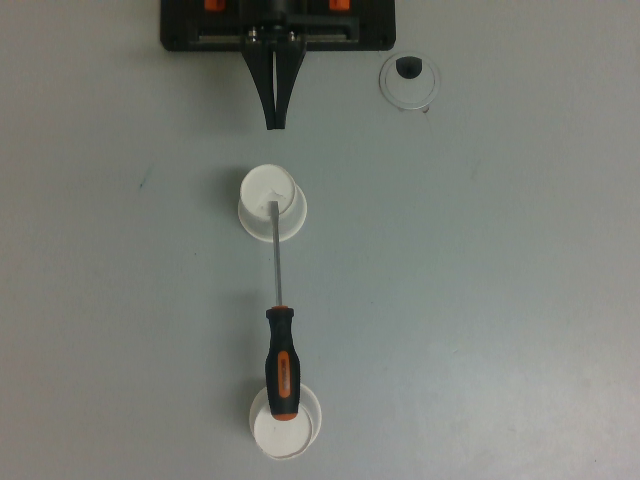}
	\end{minipage}
	\begin{minipage}{0.32\linewidth}
		\centering
		\includegraphics[width=0.9\linewidth]{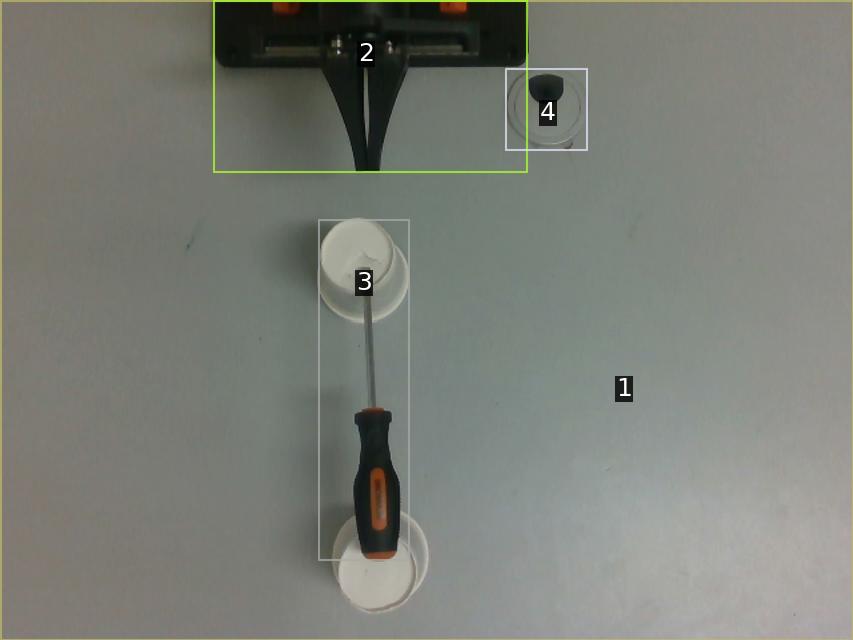}
	\end{minipage}
        \begin{minipage}{0.32\linewidth}
            \centering
		\includegraphics[width=0.9\linewidth]{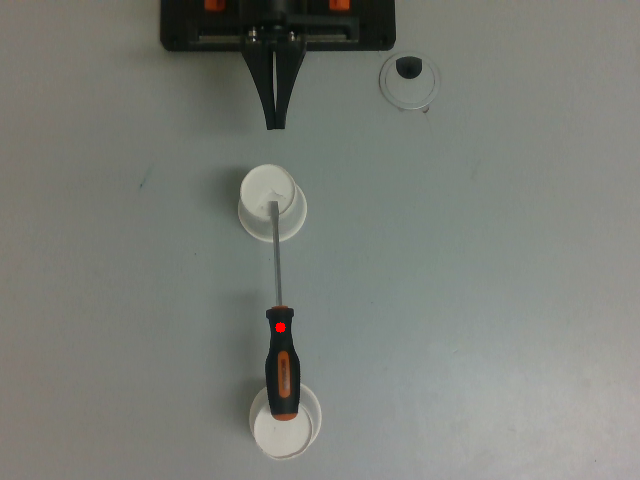}
	\end{minipage}
        \begin{minipage}{0.32\linewidth}
		\centering
		\includegraphics[width=0.9\linewidth]{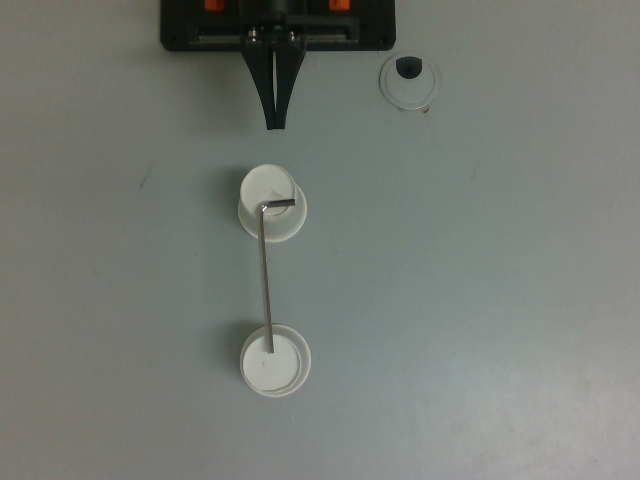}
	\end{minipage}
	\begin{minipage}{0.32\linewidth}
		\centering
		\includegraphics[width=0.9\linewidth]{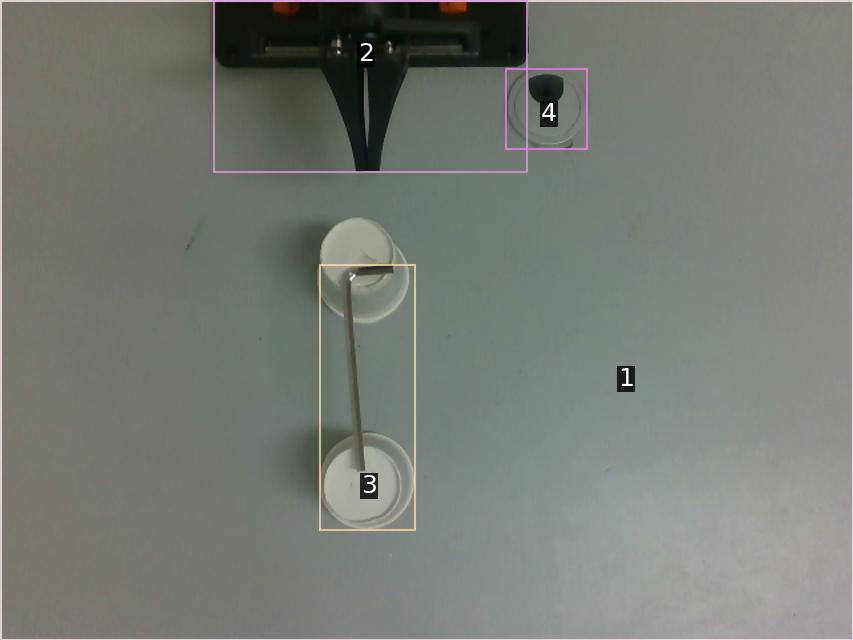}
	\end{minipage}
        \begin{minipage}{0.32\linewidth}
            \centering
		\includegraphics[width=0.9\linewidth]{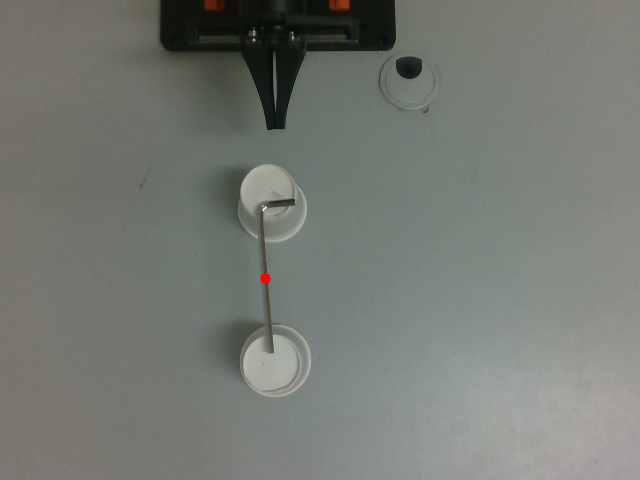}
        \end{minipage}

        \begin{minipage}{0.32\linewidth}
		\centering
		\includegraphics[width=0.9\linewidth]{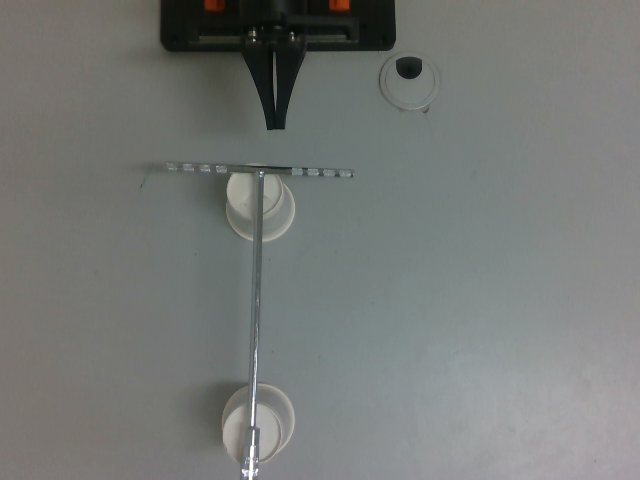}
	\end{minipage}
	\begin{minipage}{0.32\linewidth}
		\centering
		\includegraphics[width=0.9\linewidth]{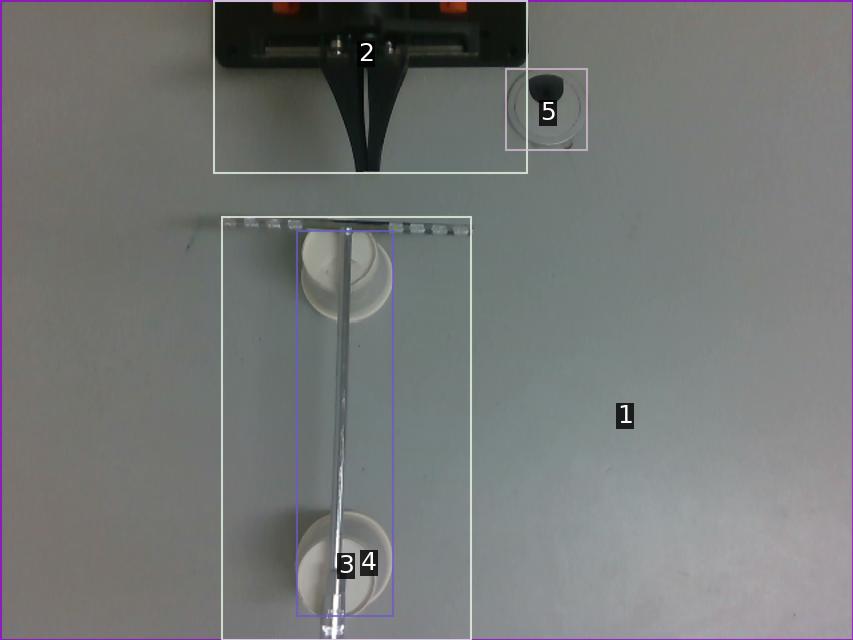}
	\end{minipage}
        \begin{minipage}{0.32\linewidth}
            \centering
		\includegraphics[width=0.9\linewidth]{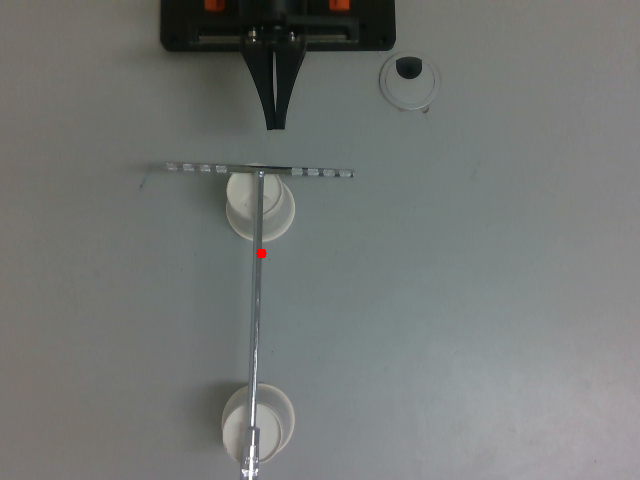}
	\end{minipage}
\end{figure}

\end{document}